\documentclass[preprint,12pt]{elsarticle}

\usepackage{amssymb}
\usepackage{amsmath}

\usepackage{graphicx}%
\usepackage{multirow}%
\usepackage{amsmath,amssymb,amsfonts}%
\usepackage{amsthm}%
\usepackage{mathrsfs}%
\usepackage[title]{appendix}%
\usepackage{xcolor}%
\usepackage{textcomp}%
\usepackage{manyfoot}%
\usepackage{booktabs}%
\usepackage{algorithm}%
\usepackage{algorithmicx}%
\usepackage{algpseudocode}%
\usepackage{listings}%
\usepackage{booktabs}    	
\usepackage{bbding} 
\usepackage[capitalize]{cleveref}
\usepackage{subcaption}
\usepackage{array}
\usepackage{pifont}
\usepackage{soul, color, xcolor}
\usepackage{natbib}

\journal{Nuclear Physics B}

\begin{document}

\begin{frontmatter}



\title{FMDConv: Fast Multi-Attention Dynamic Convolution via Speed-Accuracy Trade-off} 

\author[label1]{Tianyu Zhang}
\author[label1]{Fan Wan}
\author[label1]{Haoran Duan}
\author[label2]{Kevin W. Tong}
\author[label1]{Jingjing Deng}
\author[label1]{Yang Long\corref{cor1}}
\ead{yang.long@ieee.org}

\cortext[cor1]{Corresponding author}

\affiliation[label1]{organization={Computer Science Department, Durham University},
            addressline={The Palatine Centre, University, Stockton Rd}, 
            city={Durham},
            postcode={DH1 3LE}, 
            state={County Durham},
            country={United Kingdom}}

\affiliation[label2]{organization={College of Automation \& College of Artificial Intelligence, Nanjing University of Posts and Telecommunications},
            addressline={Guangdong Rd, Gulou}, 
            city={Nanjing},
            postcode={210023}, 
            state={Jiangsu},
            country={China}}

\begin{abstract}
Spatial convolution is fundamental in constructing deep Convolutional Neural Networks (CNNs) for visual recognition. While dynamic convolution enhances model accuracy by adaptively combining static kernels, it incurs significant computational overhead, limiting its deployment in resource-constrained environments such as federated edge computing. To address this, we propose Fast Multi-Attention Dynamic Convolution (FMDConv), which integrates input attention, temperature-degraded kernel attention, and output attention to optimize the speed-accuracy trade-off. FMDConv achieves a better balance between accuracy and efficiency by selectively enhancing feature extraction with lower complexity. Furthermore, we introduce two novel quantitative metrics, the Inverse Efficiency Score and Rate-Correct Score, to systematically evaluate this trade-off. Extensive experiments on CIFAR-10, CIFAR-100, and ImageNet demonstrate that FMDConv reduces the computational cost by up to 49.8\% on ResNet-18 and 42.2\% on ResNet-50 compared to prior multi-attention dynamic convolution methods while maintaining competitive accuracy. These advantages make FMDConv highly suitable for real-world, resource-constrained applications.
\end{abstract}







\end{frontmatter}


\section{Introduction}\label{sec1}

Convolutional Neural Networks (CNNs) \cite{liu2024bidirectional, yang2024optimal, zhou2022set, chen2023mask} have become the dominant approach for various vision-based tasks, including object detection \cite{amit2020object, hoanh2024focus}, semantic segmentation \cite{guo2018review, yue2024cross}, and image classification \cite{rawat2017deep, rahman2024activation}. While traditional networks like VGGNets, GoogLeNets, and ResNets rely on static convolutional kernels, these fixed-size kernels limit the ability to capture diverse contexts in images with varying scales and resolutions.

To address this, recent studies have explored dynamic convolution \cite{huang2024domain, duan2023dynamic}, where kernels adapt to different input characteristics. SENet \cite{hu2018squeeze} introduced dynamic channel weighting, while CondConv \cite{yang2019condconv} further extended dynamic convolution by constructing unique kernels for individual images. DynamicConv \cite{chen2020dynamic} and ODConv \cite{li2022omni} incorporated attention mechanisms into dynamic convolution, significantly improving the adaptability of convolutional operations and enhancing feature representation learning.

However, despite these advancements, existing methods struggle to effectively balance the trade-off between computational efficiency and accuracy. Many prior works focus primarily on improving accuracy but overlook the need for efficiency in real-world applications, particularly in edge computing and mobile environments. While ODConv \cite{li2022omni} integrates multiple attention mechanisms to enhance feature extraction, its complexity leads to substantial computational cost increases, making it less feasible for deployment in resource-limited scenarios.

A key challenge in deep learning is the speed-accuracy trade-off, which remains inadequately addressed. Existing studies primarily rely on empirical observations, using FLOPs or inference time as proxies for efficiency, but struggle to provide a systematic framework to jointly evaluate both efficiency and accuracy. The lack of quantifiable measures limits the development of efficiency-balanced deep learning models for broader applications. Moreover, most prior studies assess trade-offs based on empirical observations rather than theoretical formulation, leading to inconsistencies in evaluating efficiency-performance balance. The speed-accuracy trade-off, a well-established concept in psychology and neuroscience \cite{reed1973speed}, offers valuable insights for computational science.

\begin{figure}[htbp]
\centering
\includegraphics[width=0.8\textwidth]{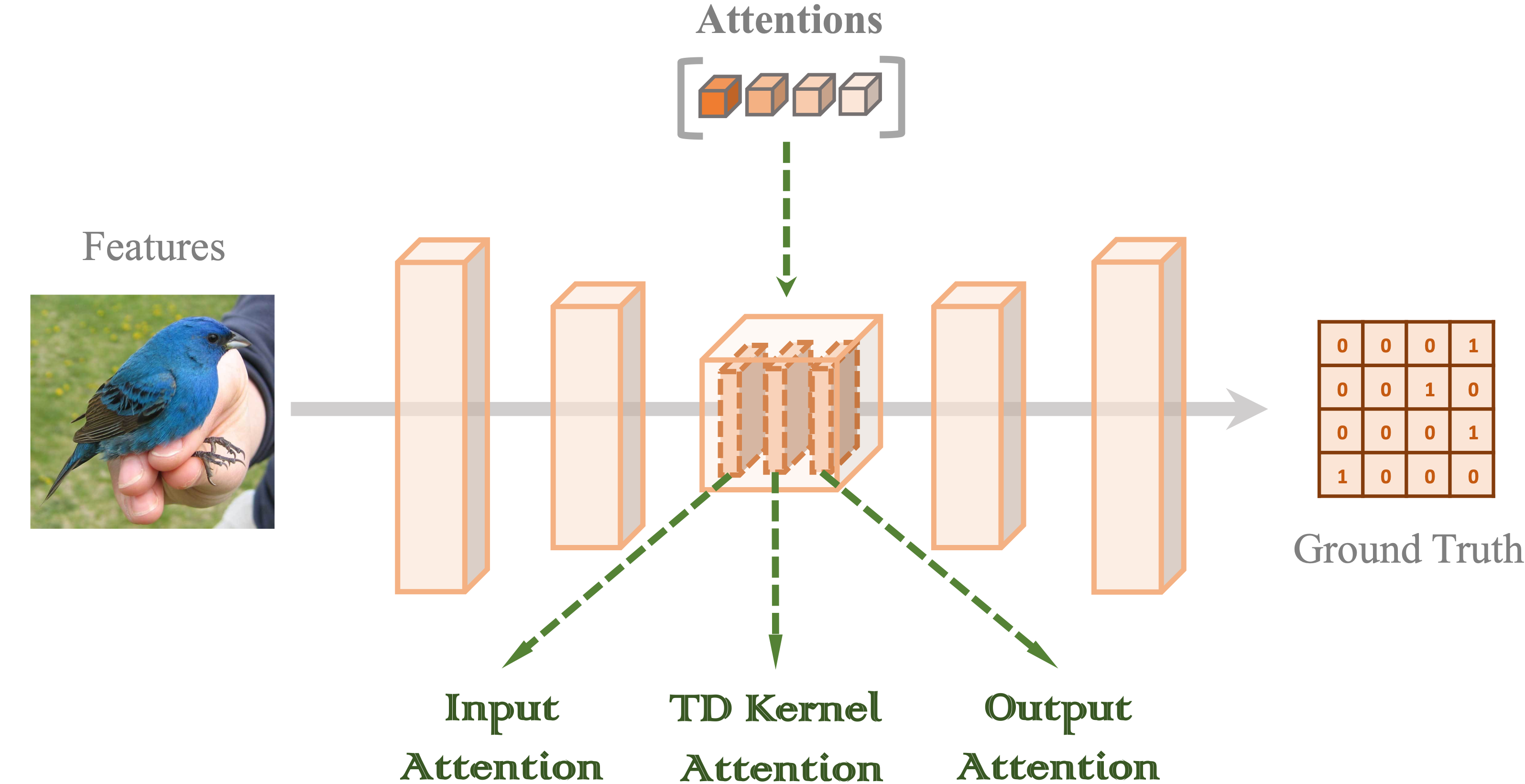}
\caption{Overview of the FMDConv framework. The diagram illustrates the three attention mechanisms (Input, TD Kernel, and Output Attention) in FMDConv, each targeting a distinct stage of feature extraction for optimal efficiency and accuracy.}\label{Figure: Figure 1}
\end{figure}

To address these limitations, we introduce Fast Multi-Attention Dynamic Convolution (FMDConv), as shown in Figure \ref{Figure: Figure 1}, a novel lightweight convolutional block that selectively integrates multiple attention mechanisms—input attention, temperature-degraded kernel attention, and output attention—to reduce computational complexity while maintaining competitive accuracy.

Additionally, we propose two novel quantitative metrics, the Inverse Efficiency Score (IES) and the Rate-Correct Score (RCS), to systematically evaluate the efficiency-accuracy trade-off in deep learning architectures. Unlike prior works that only measure FLOPs or inference time, our metrics provide a unified framework to jointly assess computational efficiency and model accuracy, enabling standardized comparisons across different approaches.

Extensive experiments on CIFAR-10, CIFAR-100, and ImageNet demonstrate that FMDConv achieves state-of-the-art efficiency-accuracy trade-offs, reducing FLOPs by up to 49.8\% on ResNet-18 and 42.2\% on ResNet-50, while maintaining competitive accuracy. Our findings highlight FMDConv’s potential for real-world, resource-constrained applications and underscore the necessity of standardized efficiency-performance evaluations in deep learning.

Our main contributions can be summarized as follows:
\begin{itemize}
\setlength\itemsep{0em}
\item We propose IES \& RCS, the first standardized metrics for evaluating the speed-accuracy trade-off in dynamic convolution, moving beyond previous qualitative assessments.
\item We conduct a comprehensive evaluation of attention mechanisms in dynamic convolution and identify input attention, temperature-degraded kernel attention, and output attention as the optimal structures for balancing efficiency and accuracy.
\item We introduce Fast Multi-Attention Dynamic Convolution (FMDConv), a novel lightweight convolutional block that selectively integrates these attentions, significantly reducing computational cost while maintaining accuracy. Compared to ODConv, FMDConv achieves up to 49.8\% and 42.2\% FLOP reductions on ResNet-18 and ResNet-50, respectively, making it highly efficient for resource-constrained applications.
\item Extensive experiments on CIFAR-10, CIFAR-100, and ImageNet demonstrate the superiority of FMDConv, achieving a state-of-the-art efficiency-accuracy trade-off with reduced computational overhead. 
\end{itemize}

\section{Related Works}\label{sec2}
\textbf{Deep Convolutional Neural Networks (CNNs) Architecture}.
Deep convolutional neural networks (CNNs) have been widely employed in computer vision \cite{wan2024sentinel, gao2023privacy, wan2023community, li2024sid, duan2025parameter, miao2025laser, long2017zero, ding2021repvgg} due to their ability to hierarchically extract feature representations from raw visual data. LeNet-5 is one of the first convolutional neural networks proposed by LeCun \emph{et. al.} \cite{lecun1998gradient} in the late 1990s, which was trained using backpropagation for a handwritten zip code recognition task with superior performance.
AlexNet, invented by Krizhevsky \emph{et. al.,} won the ImageNet 2012 challenge and significantly outperformed the second place by over 10\% in both classification and localization tasks. It adopts convolutional layers, dropout regularization, and data augmentation strategies, and its success demonstrates the great potential of deep CNNs in visual recognition. By then, limited by computational resources, their work did not explore deeper architecture. Simonyan \emph{et. al.} from Oxford proposed a family of VGGNets \cite{simonyan2014very}, which introduces a systematic approach to designing deeper models. For example, the paper proposes using stacked small filters (\emph{i.e.,} 2 stacked 3x3 kernels) to replace a single large filter (\emph{i.e.,} 1 single 5x5 kernel), which can span the same receptive field with a deeper architecture but fewer computational operations. They showed that the number of layers positively correlates with the model accuracy of up to 16 layers, which outperforms AlexNet by a significant margin. However, they also observed that adding more layers (\emph{i.e.,} 19) cannot further improve the performance. 
InceptionNet \cite{szegedy2015going}, also known as GoogLeNet, takes a novel approach to expand the architecture horizontally within the same layer by using multiple kernels at different scales, where the computed feature maps are then concatenated to form the input for the next layer. It also uses numerous auxiliary classifiers at different levels to improve training convergence. He \emph{et. al.} \cite{he2016deep} proposed a ResNet that uses residual connection structure to learn the depth of the architecture dynamically via backpropagation. ResNet won several visual recognition challenges and was used as the core for AlphaGo. 

Lightweight architecture design has attracted great attention.
Howard \emph{et. al.} \cite{howard2017mobilenets} introduced MobileNet, which is suitable for deploying on mobile and embedded devices. The model decomposes a standard convolutional filter into a depth-wise convolution and a point-wise convolution, greatly reducing computational operations. In addition, group convolution and channel shuffling used in ShuffleNet \cite{zhang2018shufflenet} are often used jointly with other design principles. In this paper, we consider both efficiency and performance from a trade-off perspective, and the proposed FMDConv can be used as a basic building block or a design paradigm in convolutional neural network architecture. These backbone networks, as we mentioned above, are mainly based on static convolution operators and have been widely adopted in many subsequent deep CNNs, which has stimulated further research. 

\noindent\textbf{Dynamic Convolution Neural Networks}.
The core idea of dynamic convolution originated from ConvCond, proposed by Yang \emph{et. al.} \cite{yang2019condconv} in 2019. A static convolution applies the same kernel across the whole dataset; a dynamic convolution has a unique kernel for each image, which can be achieved by using parameterized convolutions that are conditioned on the input images. Chen \emph{et. al.} \cite{chen2020dynamic} proposed to use the attention mechanism over the kernel itself, which dynamically integrates multiple parallel convolution kernels into one that is conditioned on the layer input. The experimental results showed that such an integration strategy can improve expression capacity without increasing the depth and width of the network. We refer the readers to the works of Han et al. \cite{han2021dynamic}, Sun et al. \cite{sun2022object}, Wu et al. \cite{wu2022attcluster}, Zhao et al. \cite{zhao2022mask}, Huang et al. \cite{huang2022image}, and Zhang et al. \cite{zhang20193d} for existing works and open problems in constructing dynamic convolution. It is worth noting that most works endow convolution kernels with dynamic properties through the single dimension of the kernel space (\emph{i.e.,} the number of convolution kernels). Li \emph{et. al.} \cite{li2022omni} argued that integrating all dimensions (including kernel attention, output channel attention, input channel attention, and spatial attention) is capable of improving the learning capacity of the model and achieving better recognition performance. However, dynamic convolution usually involves additional operations that are computationally expensive. The DCD Network proposed by Li \emph{et. al.} \cite{li2021revisiting} replaces dynamic attention over channel groups with channel fusion in a low-dimensional space, which requires fewer parameters and lower computational costs without sacrificing model accuracy. Inspired by ODConv \cite{li2022omni} and DynamicConv \cite{chen2020dynamic} models, in this paper, we focus on trading off efficiency and accuracy via optimizing the integration of dynamic strategies through all possible dimensions of the kernel space.

\noindent\textbf{Speed-Accuracy Trade-off}. 
Huang \cite{huang2017speed} demonstrated various feasible approaches to trading accuracy for speed and memory usage in deep learning-based object detection frameworks. Similar works were proposed for different application domains, such as Riel \emph{et. al.} \cite{castro2021assessing} for medical applications using axial Computed Tomography (CT) images, Javadi \emph{et. al.} \cite{javadi2018humanoid} for humanoid robots, and Chaves \emph{et. al.} \cite{chaves2019improving} for forensic surveillance. In summary, the speed-accuracy trade-off can be achieved via either a model pruning strategy (such as partial sequential pruning \cite{li2019partial}) or specifically designed building blocks (such as COSFORMER \cite{qin2022cosformer}, GSoP \cite{gao2019global}). We notice that the speed-accuracy trade-off \cite{reed1973speed} is a well-established concept in psychology and neuroscience that refers to the tendency of individuals to balance the speed and accuracy of their responses in a given task. Motivated by \cite{reed1973speed}, in this paper we propose two new metrics to measure the trade-off between speed and accuracy and develop FMDConv, a novel efficiency-accuracy-balanced building block for deep CNNs. 

Table~\ref{tab: Table 1} provides a structured comparison of existing static and dynamic convolution methods, emphasizing computational complexity, key mechanisms, and practical applications.

\begin{table}[thb]\centering
    \resizebox{1.0\textwidth}{!}{
    \large
    \begin{tabular}{l l l l}
        \toprule
        \textbf{Method} & \textbf{Key Technique} & \textbf{Application Scenarios} & \textbf{Limitations} \\ 
        \midrule
        VGG \cite{simonyan2014very} & Stacked small convolutional kernels & General-purpose CNNs & High computational cost \\  
        ResNet \cite{he2016deep} & Residual connections for deeper networks & Deep CNN architectures & Performance saturates at extreme depth \\  
        SENet \cite{hu2018squeeze} & Channel recalibration via SE module & Lightweight models & Does not learn spatial dependencies \\ 
        CondConv \cite{yang2019condconv} & Mixture of convolutional kernels & Efficient deep models & High memory overhead \\  
        DynamicConv \cite{chen2020dynamic} & SoftMax-weighted kernel aggregation & Image classification & High computational cost \\  
        ODConv \cite{li2022omni} & Kernel, spatial, channel, and filter attention & Large-scale vision models & Higher computational cost \\  
        \bottomrule
    \end{tabular}
    }
    \caption{Comparison of existing static and dynamic convolution methods, summarizing key techniques, application scenarios, and limitations.}
    \label{tab: Table 1}
\end{table}

\section{Methodology}\label{sec3}
In this section, we provide a detailed description of the architecture and implementation of Fast Multi-Attention Dynamic Convolution (FMDConv). FMDConv introduces multiple attention mechanisms to enhance the efficiency and accuracy of convolutional neural networks by dynamically adjusting convolutional kernel weights as well as the input and output feature attentions.

\subsection{Dynamic Convolution \& Omni-Dynamic Convolution}

Traditional static convolution applies the same kernel across all input images, whereas dynamic convolution adjusts kernel parameters dynamically according to the input image. In essence, the convolutional kernel is a learned function conditioned on the input data. Mathematically, dynamic convolution can be defined as:
\begin{equation}
    y  =  (\alpha_{w_1}^x W_1 + ... + \alpha_{w_i}^x W_i + ... + \alpha_{w_n}^x W_n) * x
\end{equation}
where $W_i$ is the weight of the $i$-th convolutional kernel, $\alpha_{w_i}^x$ is the corresponding attention value based on input $x$, $y$ represents the output feature map, and $*$ denotes the convolution operation.

Li \emph{et. al.} proposed to jointly use four different attentions in ODConv \cite{li2022omni} that can be formally defined as follows:
\begin{equation}
    y  =  (\alpha_{w_1}^x \odot \alpha_{f_1}^x \odot \alpha_{c_1}^x \odot \alpha_{s_1}^x + ... + \alpha_{w_i}^x \odot \alpha_{f_i}^x \odot \alpha_{c_i}^x \odot \alpha_{s_i}^x + ...  + \alpha_{w_n}^x \odot \alpha_{f_n}^x \odot \alpha_{c_n}^x \odot \alpha_{s_n}^x) * x
\end{equation}
where $\odot$ denotes the element-wise Hadamard product. Here, $\alpha_{w_i}^x$, $\alpha_{f_i}^x$, $\alpha_{c_i}^x$, $\alpha_{s_i}^x$ represent the kernel attention, output channel attention, input channel attention, and spatial attention, respectively, while $x$, and $y$ denote the input and output feature maps, respectively.

\subsection{Metric of Speed-Accuracy Trade-off}

Motivated by the well-established speed-accuracy trade-off concept in psychology \cite{reed1973speed}, we introduce two novel metrics, Inverse Efficiency Score (IES) \cite{townsend1983stochastic} and Rate-Correct Score (RCS) \cite{woltz2006availability}, to jointly measure computational overhead and model accuracy. Note that in cognitive psychology, reaction time (RT) refers to human subjects’ response speed. Here, we adapt the concept to measure the computational overhead (e.g., training time, FLOPs) of deep learning models. We do not imply an exact one-to-one mapping but merely draw inspiration from the speed-accuracy trade-off phenomenon.

\noindent\textbf{Inverse Efficiency Score (IES)}.
The most commonly used measure for a speed-accuracy trade-off in experimental psychology is IES \cite{townsend1983stochastic}, which is typically defined as the mean correct reaction time (RT) divided by the proportion of correct classifications. 
In our case, we adopt the concept of the original IES and formulate the score as the ratio of the training time of an epoch to the Top-1 accuracy rate:
\begin{equation}
IES_{ij} = \frac{\overline{RT_{ij}}}{PC_{ij}}
\end{equation}
where $\overline{RT_{ij}}$ is the mean training time of the model $i$ on correct-classification trials with hyper-parameter set $j$, and $PC_{ij}$ is the proportion of correct classifications of $i$ for $j$. Although most (if not all) research using IES has only included $RT$s with correct trails, the original study suggests all $RT$s (including error trials) should be taken into account.

\noindent\textbf{Rate-Correct Score (RCS)}.
We also adopt an alternative speed-accuracy trade-off metric, the rate-correct score (RCS), that can be defined as:
\begin{equation}
RCS_{ij} = \frac{NC_{ij}}{\sum_{k=1}^{n_{ij}} RT_{ijk}}
\end{equation}
where $NC_{ij}$ is the number of correct classifications of the model $i$ in condition of $j$, and the denominator reflects the total time the model $i$ spent on training in condition of $j$ (\emph{i.e.,} the sum of RTs across all $n_{ij}$ training of the model $i$ in condition of $j$). RCS can be interpreted directly as the number of correct classifications per unit of time.
Both methods compare accuracy and training time. The difference is that IES is only related to accuracy and running time, while RCS also considers the size of the training database.

\subsection{Fast Multi-Attention Dynamic Convolution}

In this section, we present the architecture of Fast Multi-Attention Dynamic Convolution (FMDConv). Unlike DynamicConv and ODConv, we introduce a multi-attention mechanism and optimize the integration of dynamic strategies across multiple kernel dimensions to enhance both efficiency and accuracy.

\subsubsection{Architecture Design}

\begin{figure}[htbp]
\centering
\includegraphics[width=0.98\textwidth]{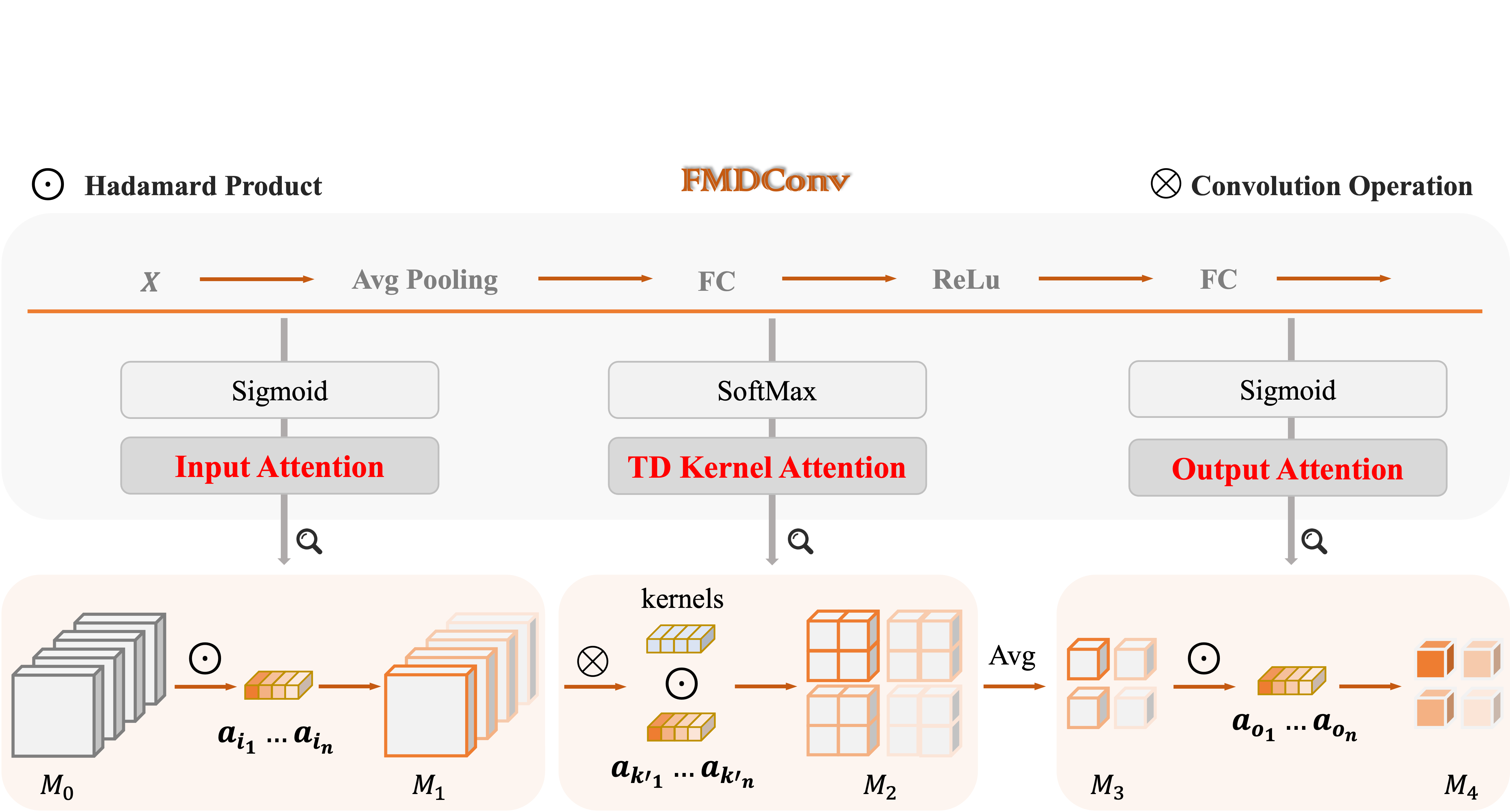}
\caption{The architecture of the Fast Multi-Attention Dynamic Convolution (FMDConv) block. It integrates three attention mechanisms: Input Attention, Temperature-Degraded (TD) Kernel Attention, and Output Attention. These attentions are computed via Sigmoid and SoftMax functions to adjust feature maps and convolution kernels dynamically.}\label{Figure: Figure 2}
\end{figure}

In this subsection, we present the proposed FMDConv block, namely Fast Multi-Attention Dynamic Convolution. Similar to Dynamic Convolution \cite{chen2020dynamic} and Omni-Dynamic Convolution \cite{li2022omni}, we adopt a multi-attention mechanism and calculate convolution from N learnable kernels with the same spatial size and channel dimension. However, based on those two proposed speed-accuracy trade-off metrics, we conclude that calculating kernel attention and spatial attention are computationally expensive, while both contribute very little to improving accuracy. Therefore, we replace these two attention mechanisms with temperature-degraded kernel attention originating from DynamicConv \cite{chen2020dynamic}. The overall architecture of the proposed FMDConv is illustrated in Figure \ref{Figure: Figure 2}.
The proposed FMDConv can be formulated as:
\begin{equation}
    y  =  (\alpha_{i_1} \odot \alpha_{k'_1} \odot \alpha_{o_1} + \alpha_{i_2} \odot \alpha_{k'_2} \odot \alpha_{o_2} + ... + \alpha_{i_n} \odot \alpha_{k'_n} \odot \alpha_{o_n}) * x
\end{equation}
where $x$, and $y$ denote the input and output feature maps, respectively, while $\alpha_{i_n}$, $\alpha_{k'_n}$ and $\alpha_{o_n}$ correspond to the input channel attention, temperature-degraded kernel attention, and output channel attention, respectively, as detailed in Algorithm \ref{FMDConv}.

As shown in Figure \ref{Figure: Figure 2}, in input attention and output attention, the input information will first be squeezed by global average pooling, followed by a fully connected layer, a ReLu activation layer, and another fully connected layer to calculate the sample-dependent information. To compute the final attention values \{$\alpha_c, \alpha_k', \alpha_f$\}, we again apply different non-linear activation operators individually on the extracted sample-dependent feature for input attention, temperature-degraded kernel attention, and output attention. For input and output attention activation functions, sigmoid is used, while we use SoftMax for kernel attention instead.

\begin{algorithm}
\caption{Fast Multi-Attention Dynamic Convolution (FMDConv2d)}
\label{FMDConv}
\begin{algorithmic}[1]
\State \textbf{Input:} Input tensor $x$, temperature $T$, kernel size $k$, number of kernels $K$, input channels $C_{\text{in}}$, output channels $C_{\text{out}}$
\State \textbf{Output:} Output tensor after dynamic convolution
\State
\State \textbf{Initialization:}
\State Initialize attention layers $\text{Attention}$, $\text{Attention2}$, and convolution kernels $W$ with Kaiming initialization.
\State Initialize bias terms $b$ if applicable.
\State
\State \textbf{Step 1: Input Attention Computation}
\State Compute input attention $A_{\text{input}} = \sigma(\text{Conv}(x))$  \Comment{$\sigma$: Sigmoid activation}
\State Compute output attention $A_{\text{output}} = \sigma(\text{Conv}(x))$
\State 
\State \textbf{Step 2: Kernel Attention Computation}
\State Compute kernel attention $A_{\text{kernel}} = \text{Softmax}\left(\frac{\text{Attention2}(x)}{T}\right)$ \Comment{Temperature-scaled softmax}
\State 
\State \textbf{Step 3: Dynamic Convolution Calculation}
\State Multiply input $x$ by input attention: $x = x \times A_{\text{input}}$
\State Reshape input tensor $x$ to shape $(1, C_{\text{in}} \times \text{batch size}, H, W)$
\State Reshape convolution kernel weights $W$ to shape $(K, C_{\text{out}} \times C_{\text{in}}, k, k)$
\State Compute aggregate weight matrix: $W_{\text{agg}} = A_{\text{kernel}} \times W$
\If{bias is not None}
    \State Compute aggregate bias $b_{\text{agg}} = A_{\text{kernel}} \times b$
    \State Perform convolution: $\text{Output} = \text{Conv2d}(x, W_{\text{agg}}, b_{\text{agg}})$
\Else
    \State Perform convolution: $\text{Output} = \text{Conv2d}(x, W_{\text{agg}})$
\EndIf
\State Reshape output to batch size and apply output attention: $\text{Output} = \text{Output} \times A_{\text{output}}$
\State
\State \textbf{Return:} Output tensor after applying multi-attention dynamic convolution.
\end{algorithmic}
\end{algorithm}

\subsubsection{Temperature of SoftMax Activation Function}
The SoftMax activation function is commonly used for kernel attention, but in the early stages of training, the uniform output of SoftMax can lead to slow convergence. To address this, we introduce a temperature degradation mechanism where the initial temperature is set to 40 and decreases by 3 after each epoch until the temperature reaches 1. The formula for temperature-degraded SoftMax is given by:
\begin{equation}
    \sigma(z_i) = \frac{e^{z_{i}/T}}{\sum_{j=1}^K e^{z_{j}/T}} \ \ \
\end{equation}
where $\sigma(z_i)$ denotes the output probability for the $i$-th kernel, $z_i$ is the $i$-th element of the input vector $z$, $K$ is the total number of kernels, and $T$ is the temperature parameter that controls the sharpness of the SoftMax distribution. We set the initial temperature to start at 40 and subtract 3 after each epoch of training until T. When $T=1$, the formula is identical to normal SoftMax. 
Setting the temperature to decrease by 3 every epoch can greatly mitigate the slow start issue of SoftMax at the early epoch, which produces near one-hot output at the beginning of the training. When the temperature T = 1, this function reduces to the standard SoftMax. Our experiments show that setting the initial temperature to 40 improves Top-1 accuracy by 2.95\% on the CIFAR-100 dataset.

\section{Experiments}\label{sec4}
\subsection{Benchmark and Experiment Setting}
To evaluate the proposed FMDConv, we adopted ResNet-18 as the backbone due to its low computational and memory requirements, making it suitable for resource-constrained platforms. The experiments were conducted on widely used benchmarks, including CIFAR-10, CIFAR-100, and ImageNet (ILSVRC 2012). The CIFAR datasets consist of 50,000 training images and 10,000 testing images, while the ILSVRC2012 dataset contains 1,281,167 training images and 50,000 validation images across 1000 categories. Compared to CIFAR datasets, ImageNet offers higher resolution, and more diverse image categories, and presents a greater challenge for classification tasks.

In this study, we initially tuned the hyperparameters based on the training set performance, as our model demonstrated robustness across a wide range of hyperparameter settings (as presented in Section 4.4). However, following the reviewer's suggestion and adhering to best practices, we recognize the importance of using a validation set for hyperparameter optimization. In future work, we plan to incorporate a separate validation set to ensure hyperparameter tuning is conducted independently of the test set, thereby further enhancing the robustness and generalizability of our results.
 
For comparative analysis, we evaluated our method against existing dynamic convolution approaches, such as ODConv \cite{li2022omni}, DynamicConv, and CondConv, to comprehensively demonstrate the advantages of FMDConv.

In our training regimen for both CIFAR-10 and CIFAR-100, the initial learning rate was set to 0.1, with a decay factor of 20 applied every 30 epochs over 70 epochs of training. We used a weight decay of 1e-4, a dropout rate of 0.1, and a reduction rate of 0.0625. The training batch size was 32, and the test batch size was 70. For ImageNet, we adopted a different strategy, initializing the learning rate at 0.1 and reducing it by a factor of 30 every 30 epochs, for a total of 100 epochs.

All experiments were conducted on a system with an NVIDIA GeForce RTX 3080 GPU (10GB GDDR6X memory), 32GB Corsair VENGEANCE RGB PRO DDR4 RAM, and an Intel® Core i9-12900K CPU. The software environment included PyTorch 1.12.1, CUDA 11.3, and Python 3.9.

\subsection{Speed-Accuracy Trade-off Evaluation}
We first evaluate the speed-accuracy trade-off of four attention mechanisms used in Omni-Dynamic Convolution with the proposed metrics (IES and RCS) on an image classification task. Table \ref{tab: Table 2} illustrates the accuracy and time consumption of channel attention, kernel attention, spatial attention, and filter attention, respectively, on the CIFAR-10 dataset. The best two results are highlighted in bold. Our findings reveal that channel attention and filter attention achieve better accuracy (improvements of 1.13\% and 1.12\% on Top-1 accuracy, respectively) with a relatively small increase (12.53 seconds and 13.00 seconds extra per training epoch) in time consumption. However, kernel attention and spatial attention lead to significant time consumption with limited improvement in accuracy. We further conducted experiments on kernel attention with two and four kernels, where the kernel attention improves Top-1 accuracy by 0.45\% with an extra 97.71 seconds of time cost per training epoch when the kernel number is two and by 0.9\% with an extra 158.53 seconds with four kernels. We also used RCS as the key index to measure the effectiveness of these four attention mechanisms. We found that channel attention and filter attention outperform kernel attention and spatial attention with RCS scores of 915.32 and 908.05, compared to kernel attention and spatial attention with RCS scores of 373.76 and 481.63, respectively.

\begin{table}[thb]\centering
    \resizebox{1.0\textwidth}{!}{
    \large
    \begin{tabular}{*{10}{c}}
        \toprule
       kernel Number & Channel Attention & Kernel Attention & Spatial Attention & Filter Attention & Top-1\% & Top-5\% & Time per Epoch/s & IES & RCS\\
        \midrule
        1 & - & - & - & - & 89.7 & 99.65 & 47.01 & - & - \\
        1 & \Checkmark & - & - & - & $\mathbf{90.83}$ & $\mathbf{99.68}$ & $\mathbf{59.54}$ & $\mathbf{65.65}$ & $\mathbf{915.32}$\\
        2 & - & \Checkmark & - & - & 90.15 & 99.72 & 144.72 & 160.53 & 373.76\\
        4 & - & \Checkmark & - & - & 90.6 & 99.71 & 205.54 & 226.87 & 264.47\\
        1 & - & - & \Checkmark & - & 90.41 & 99.76 & 112.63 & 124.58 & 481.63\\
        1 & - & - & - & \Checkmark & $\mathbf{90.82}$ & $\mathbf{99.82}$ & $\mathbf{60.01}$ & $\mathbf{66.08}$ & $\mathbf{908.05}$\\
        \bottomrule
    \end{tabular}}
    \caption{Comparison of results of each attention on the Cifar-10 validation set with the ResNet18 backbones trained for 100 epochs. We set r = 0.1. The best results are bold.}
    \label{tab: Table 2}
\end{table}

We conclude that spatial attention and kernel attention have little impact on the Top-1 accuracy of the CIFAR-10 dataset while significantly increasing time consumption.

\begin{figure}[htbp]
    \centering
    \begin{subfigure}[b]{0.32\linewidth}
        \centering
        \includegraphics[width=4cm]{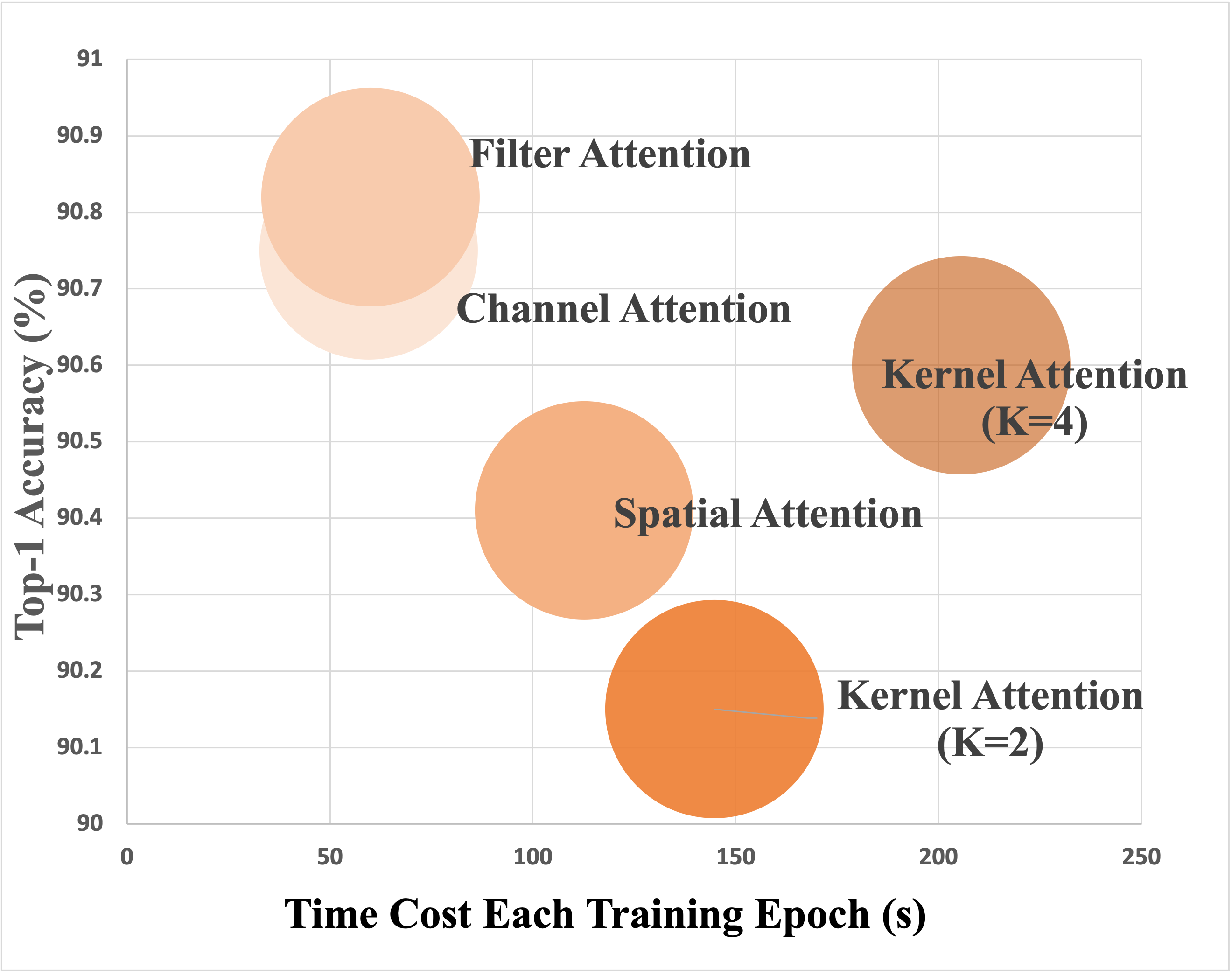} 
        \caption{Attentions Comparison}
    \end{subfigure}
    \begin{subfigure}[b]{0.32\linewidth}
        \centering
        \includegraphics[width=4cm]{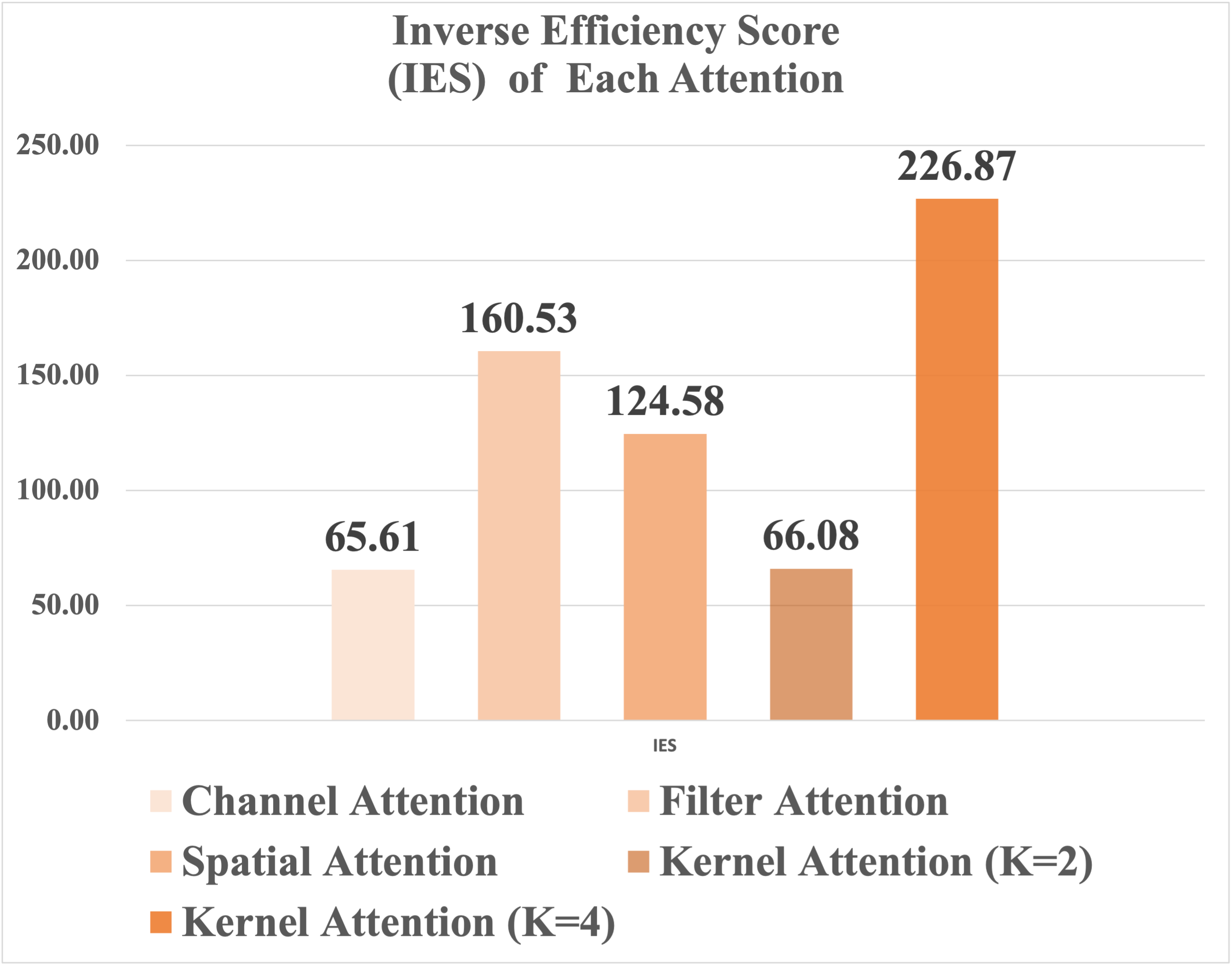}
        \caption{IES}
    \end{subfigure}
    \begin{subfigure}[b]{0.32\linewidth}
        \centering
        \includegraphics[width=4cm]{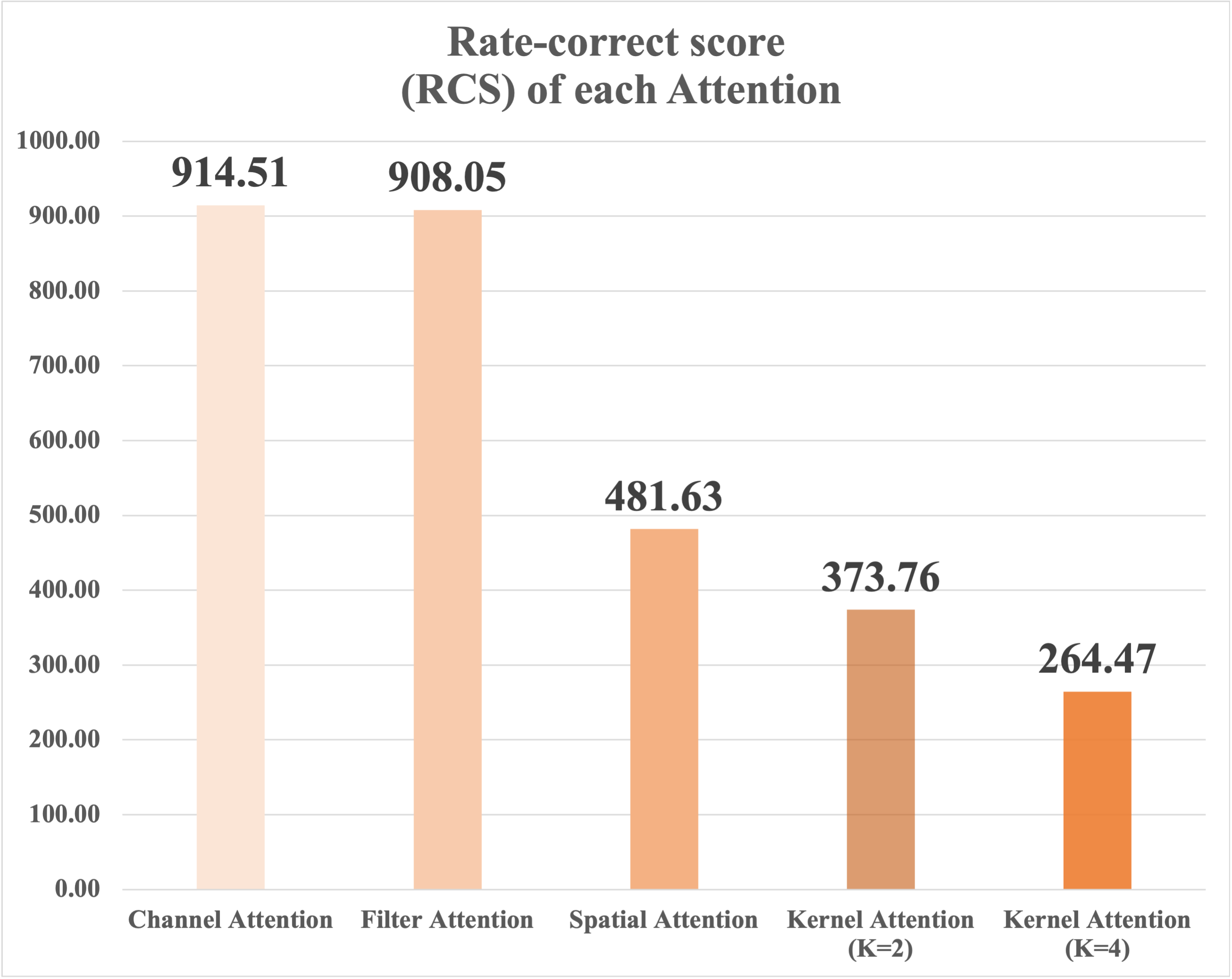}
        \caption{RCS}
    \end{subfigure}
    \caption{(a) Attentions Comparison on the CIFAR-10; (b) Inverse Efficiency Score (IES); (c) Rate-correct Score (RCS).}
    \label{Figure: Figure 3}
\end{figure}

In Figure \ref{Figure: Figure 3} (a), we present the experimental effects of the four attention mechanisms on CIFAR-10. The horizontal axis represents the time of each training epoch, and the vertical axis represents the percentage of Top-1 accuracy. The bubbles in the upper-left corner indicate higher Top-1 accuracy rates with less time and better results. Figure \ref{Figure: Figure 3} (b) illustrates the IES, which is the ratio of time to accuracy. A smaller ratio indicates better results with high accuracy and less time. Figure \ref{Figure: Figure 3} (c) displays the RCS of each attention mechanism, which is the number of correct classifications per unit of time. A higher number indicates better results, implying that more correct images can be classified within a certain period.

\subsection{Fast Multi-Attention Dynamic Convolution}

We evaluate the performance of our FMDConv model on CIFAR-10, CIFAR-100, and ImageNet. The results of our experiments, presented in \cref{tab: Table 3}, show that our FMDConv model achieves the highest Top-1 accuracy of 94.21\% and Top-5
accuracy of 99.85\%, with a time per epoch of 114.7 seconds in CIFAR-10. In addition, FMDConv outperforms the baseline regarding IES and RCS scores by a significant margin, with 121.75 in IES points and 241.08 in RCS scores.

\begin{table}[thb]\centering
    \resizebox{1.0\textwidth}{!}{
    \large
    \begin{tabular}{*{10}{c}}
        \toprule
        Model & Top-1 Accuracy & Top-5 Accuracy & Time Cost Each Epoch & IES & RCS \\ 
        \midrule
        CondConv & 81.19 & 98.96 & 100.50s & 123.78 & 484.72 \\  
        DynamicConv  & 85.19 & 99.33 & 104.05s & 122.14 & 491.24 \\ 
        ODConv & 93.82 & 99.82 & 223.61s & 238.34 & 251.74 \\
        $\mathbf{Ours}$ & $\mathbf{94.21}$& $\mathbf{99.85}$ & 114.7s & $\mathbf{121.75}$ & $\mathbf{492.82}$ &\\ 
        \bottomrule
    \end{tabular}
    }
    \caption{Comparison of results on the Cifar-10 validation set with the ResNet18 backbones trained for 70 epochs. We set r = 0.1. The best results are bold.}
    \label{tab: Table 3}                      
\end{table}

Similarly, on CIFAR-100, the FMDConv model achieves a top-1 accuracy of 74.99\% and a top-5 accuracy of 93.61\%, with a time per epoch of 115.16 seconds in CIFAR-100. \cref{tab: Table 4} illustrates that our approach achieves 153.16 in IES points and 390.71 in RCS score.

\begin{table}[thb]\centering
    \resizebox{1.0\textwidth}{!}{
    \large
    \begin{tabular}{*{10}{c}}
        \toprule
        Model & Top-1 Accuracy & Top-5 Accuracy & Time Cost Each Epoch & IES & RCS\\ 
        \midrule
        CondConv & 66.80 & 85.24 & 108.50s & 162.42 & 369.40\\
        DynamicConv & 67.21 & 86.89 & 105.98s & 157.68 & 380.50\\ 
        ODConv & 72.63 & 92.13 & 222.1s & 305.79 & 196.21\\
        $\mathbf{Ours}$  & $\mathbf{74.99}$ & $\mathbf{93.61}$ & 115.16s & $\mathbf{153.57}$ & $\mathbf{390.71}$\\ 
        \bottomrule
    \end{tabular}
    }
    \caption{Comparison of results on the Cifar-100 validation set with the ResNet18 backbones trained for 100 epochs. We set r = 0.1. The best results are bold.}
    \label{tab: Table 4}
\end{table}

In our ImageNet training experiments, when we applied identical parameters, including learning rate, batch size, and number of epochs, our approach exhibited a substantial reduction in training time, nearly halving it in comparison to ODConv. Specifically, Table \ref{tab: Table 5} compares the performance of different dynamic convolution models on the ImageNet validation set using ResNet18 as the backbone, trained for 100 epochs. Our proposed method, FMDConv ($\times$4), achieved the highest Top-1 accuracy of 73.21\% and Top-5 accuracy of 90.88\%, outperforming CondConv, DynamicConv, and ODConv. In terms of efficiency, FMDConv showed clear advantages, with a time per epoch of 620.34 seconds, which is nearly half that of ODConv (1236.14 seconds). Additionally, FMDConv achieved the lowest Inverse Efficiency Score (IES) of 847.34 and the highest Rate-Correct Score (RCS) of 1511.98, demonstrating that it offers the best trade-off between speed and accuracy. 

\begin{table}[htbp]
\resizebox{\textwidth}{!}{
\begin{tabular}{*{6}{l}}
    \toprule
    Model & Top-1 (\%) & Top-5 (\%) & Time per Epoch (s) & IES & RCS \\ 
    \midrule
    CondConv ($\times$8) & 71.99 & 90.27 & 625.68s & 869.12 & 1474.10\\
    DynamicConv ($\times$4) & 72.76 & 90.79 & 618.45s & 849.98 & 1507.28\\
    ODConv ($\times$4) & 73.09 & 90.86 & 1236.14s & 1691.26 & 757.52\\
    $\mathbf{Ours (\times 4)}$ & $\mathbf{73.21}$  & $\mathbf{90.88}$ & $\mathbf{620.34s}$ & $\mathbf{847.34}$ & $\mathbf{1511.98}$\\ 
    \bottomrule
\end{tabular}
}
\caption{Comparison of results on the ImageNet validation set with the ResNet18 backbones trained for 100 epochs. We set r = 0.0625. The best results are bold.}
\label{tab: Table 5}
\end{table}

Based on the experimental results presented in Table \ref{tab: Table 6}, we compare the performance of various dynamic convolution models on the ImageNet validation set using ResNet50 as the backbone, trained for 100 epochs. Our proposed method, FMDConv ($\times$4), achieved the best Top-1 accuracy of 78.34\% and Top-5 accuracy of 93.57\%, marginally outperforming ODConv ($\times$4) in both metrics. Notably, FMDConv significantly reduced the computational overhead compared to ODConv, with a much lower Time per Epoch of 1028.57 seconds compared to 1780.35 seconds for ODConv. Additionally, FMDConv achieved superior efficiency, reflected in the lowest Inverse Efficiency Score (IES) of 1099.25 and a higher Rate-Correct Score (RCS) of 975.79, indicating that it provides better performance per unit of time.

\begin{table}[htbp]
\resizebox{\textwidth}{!}{
\begin{tabular}{*{6}{l}}
    \toprule
    Model & Top-1 (\%) & Top-5 (\%) & Time per Epoch (s) & IES & RCS \\
    \midrule
    CondConv ($\times$8) & 75.20 & 93.12 & 990.12 & 1316.65 & 973.05\\
    DynamicConv ($\times$4) & 75.82 & 93.16 & 1008.54 & 1330.18 & 963.16\\
    ODConv ($\times$4) & 78.32 & 93.56  & 1780.35 & 2273.17 & 563.60 \\
    $\mathbf{Ours (\times 4)}$ & $\mathbf{78.34}$ & $\mathbf{93.57}$ & 1028.57 & $\mathbf{1099.25}$ & $\mathbf{975.79}$ \\
    \bottomrule
\end{tabular}
}
\caption{Comparison of results on the ImageNet validation set with the ResNet50 backbones trained for 100 epochs. We set r = 0.0625. The best results are bold.}
\label{tab: Table 6}
\end{table}

Based on the experimental results shown in Table \ref{tab: Table 7}, we evaluate the performance of various dynamic convolution approaches on the ImageNet validation set, using MobileNetV2 (×0.5) as the backbone and training for 100 epochs. Our proposed FMDConv ($\times$4) delivered the best performance with a Top-1 accuracy of 70.23\% and a Top-5 accuracy of 92.07\%, slightly surpassing ODConv ($\times$4) in both measures. Notably, FMDConv significantly improved computational efficiency, requiring only 87.37 seconds per epoch, which is a notable reduction compared to ODConv’s 119.21 seconds. Moreover, FMDConv demonstrated enhanced overall efficiency with the lowest Inverse Efficiency Score (IES) of 124.41 and the highest Rate-Correct Score (RCS) of 10298.31, reflecting its superior balance between speed and accuracy when compared to the other models.

\begin{table}[htbp]
\resizebox{\textwidth}{!}{
\begin{tabular}{*{6}{l}}
    \toprule
    Model & Top-1 (\%) & Top-5 (\%) & Time per Epoch (s) & IES & RCS \\
    \midrule
    CondConv ($\times$8) & 66.41 & 90.32 & 83.27 & 125.39 & 10217.64\\
    DynamicConv ($\times$4) & 68.75 & 91.37 & 85.62 & 124.54 & 10287.34\\
    ODConv ($\times$4) & 70.21 & 91.95  & 119.21 & 169.79 & 7545.57 \\
    $\mathbf{Ours (\times 4)}$ & $\mathbf{70.23}$ & $\mathbf{92.07}$ & 87.37 & $\mathbf{124.41}$ & $\mathbf{10298.31}$ \\
    \bottomrule
\end{tabular}
}
\caption{Comparison of results on the ImageNet validation set with the MobileNetv2(x0.5) backbones trained for 100 epochs. We set r = 0.0625. The best results are bold.}
\label{tab: Table 7}
\end{table}

\begin{figure} \centering \includegraphics[width=0.8\textwidth]{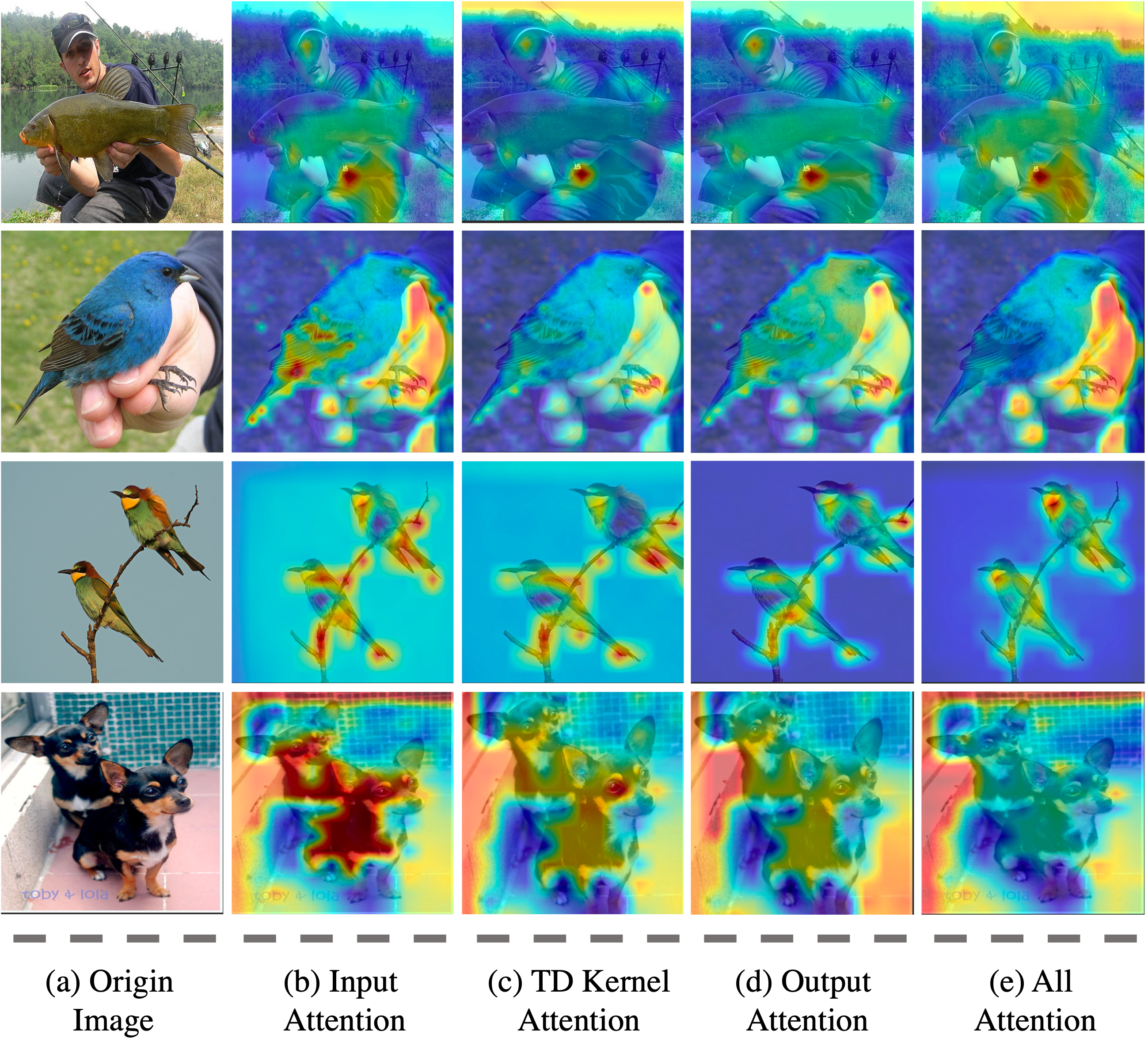} \caption{Grad-CAM++ visualization results for multiple attention mechanisms on ImageNet. (a) Original Images, (b) Feature Maps with Input Attention, (c) Feature Maps with Temperature-Degraded (TD) Kernel Attention, (d) Feature Maps with Output Attention, and (e) Feature Maps with All Combined Attentions.} \label{fig: Figure 4} \end{figure}


To further demonstrate the effectiveness of our FMDConv model, we visualize Grad-CAM++ results for different attention mechanisms using ResNet-18, as shown in Figure \ref{fig: Figure 4}. The experimental results indicate that input attention enhances the network's focus on discriminative regions before convolution operations, thereby improving the effectiveness of early feature extraction. Temperature-degraded kernel attention dynamically adjusts convolutional kernel weights, reinforcing fine-grained structural information while suppressing irrelevant background noise, leading to more stable feature representations. Output attention further refines the extracted features by emphasizing class-relevant regions and reducing redundant activations, making the final feature maps more distinct.

When all three attention mechanisms work together, the network's focus significantly improves, optimizing both spatial selectivity and class discriminability, as illustrated in Figure \ref{fig: Figure 4} (e). Unlike ODConv, which applies attention across all dimensions at a higher computational cost, FMDConv leverages a more lightweight multi-attention mechanism to achieve superior feature selectivity. In conclusion, our FMDConv model performs better in terms of model accuracy and training time cost than the baseline ODConv model on both CIFAR and ImageNet datasets. This improvement is mainly due to the introduction of the multi-attention mechanism in our FMDConv model.

\subsection{Ablation Studies}
We conducted ablation studies on the CIFAR-100 dataset to evaluate the impact of various factors on the performance of the proposed FMDConv.

\begin{figure} \centering \includegraphics[width=0.7\textwidth]{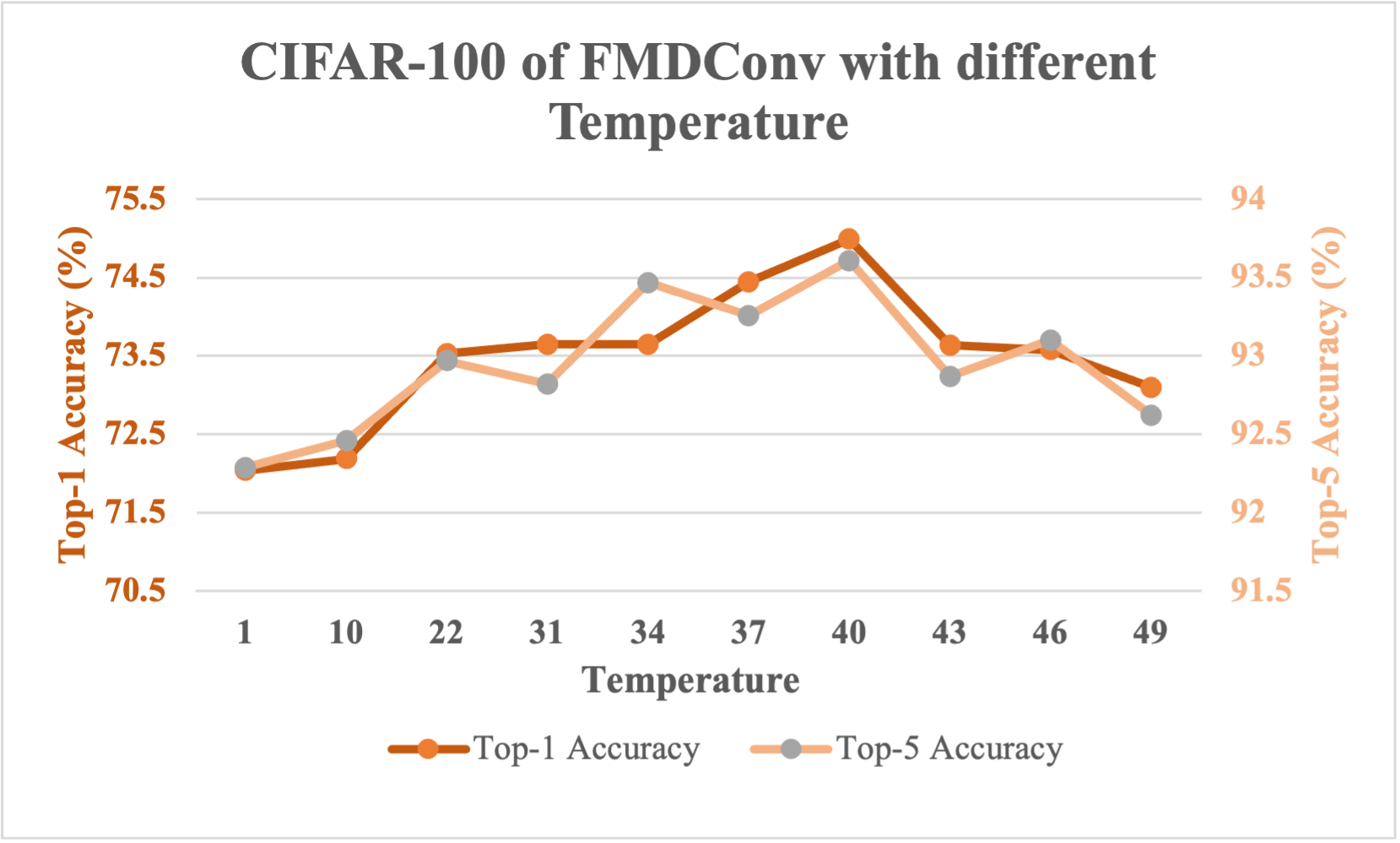} \caption{Top-1 and Top-5 accuracy comparison for FMDConv under different initial temperatures on CIFAR-100.} \label{fig: Figure 5} \end{figure}

\noindent\textbf{Effect of Initial Temperature.}
We introduced a temperature mechanism in the kernel attention module to enhance the convergence rate of dynamic convolution. As shown in Table \ref{tab: Table 8}, we compared the top-1 and top-5 accuracy rates under different initial temperatures to identify the optimal value. When the initial temperature decreased from 40, the model achieved its highest Top-1 accuracy, improving by 2.75\% compared to the baseline temperature of T = 1, as shown in Figure \ref{fig: Figure 5}. Similarly, the Top-5 accuracy increased by 1.32\%. The accuracy initially increases with temperature until it peaks at 40, after which it begins to decline. The best results are highlighted in bold.

\begin{table}[thb]\centering
\resizebox{0.6\textwidth}{!}{
\large
\begin{tabular}{*{3}{c}}
    \toprule
    \textbf{Temperature} & \textbf{Top-1 Accuracy (\%)} & \textbf{Top-5 Accuracy (\%)} \\
    \midrule
    1 & 72.04 & 92.29 \\
    10 & 72.19 & 92.46 \\
    22 & 73.53 & 92.97 \\
    31 & 73.65 & 92.82 \\
    34 & 73.65 & 93.47 \\
    37 & 74.45 & 93.26 \\
    40 & \textbf{74.99} & \textbf{93.61} \\
    43 & 73.64 & 92.87 \\
    46 & 73.58 & 93.10 \\
    49 & 73.10 & 92.62 \\
    \bottomrule
\end{tabular}}
\caption{Top-1 and Top-5 accuracy comparison of FMDConv with different initial temperatures on CIFAR-100. The best results are bold.}
\label{tab: Table 8}
\end{table}

\noindent\textbf{Effect of the Number of Convolution Kernels.}
We evaluated the impact of varying the number of convolution kernels on classification accuracy. As shown in Table \ref{tab: Table 9}, when the initial temperature is fixed at T = 40, the Top-1 accuracy reaches a maximum of 74.99\% with four kernels (K = 4). Beyond K=4, increasing the number of kernels does not lead to further improvements, as the accuracy stabilizes or decreases slightly. The corresponding Top-5 accuracy follows a similar trend, reaching 93.61\%.

\begin{table}[thb]\centering
    \resizebox{0.9\textwidth}{!}{
    \large
    \begin{tabular}{*{10}{c}}
        \toprule
        Kernel Number & Params & Top-1 Accuracy(\%) & Top-5 Accuracy(\%) & Time Cost Each Epoch(s) \\
        \midrule
        K=1 & 12.23M & 74.13 & 93.06 & 106.11 \\
        K=2 & 23.22M & 73.80 & 93.01 & 110.64\\	
        K=4 & 45.20M & $\mathbf{74.99}$ & $\mathbf{93.61}$ & 115.16\\	
        K=6 & 67.18M & 74.23 & 93.05 & 117.07\\
        K=8 & 89.16M & 74.67 & 93.09 & 120.11\\
        K=16 & 177.09M & 73.13 & 92.64 & 132.26\\
        \bottomrule
    \end{tabular}
    }
    \caption{Comparison of results of different kernel numbers on the Cifar-100 validation set with the ResNet18 backbones trained for 100 epochs. We set r = 0.1. The best results are bold.}
    \label{tab: Table 9}
\end{table}

\noindent\textbf{Effect of Learning Rate.}
In this set of experiments, we tested different initial learning rates (1/2, 1/4, 1/8, 1/10, and 1/16) to evaluate their influence on accuracy. As shown in Table \ref{tab: Table 10}, a learning rate of 1/10 produced the best Top-1 accuracy (74.99\%) and Top-5 accuracy (93.61\%) with minimal additional time per epoch. Lower learning rates did not yield significant improvements and, in some cases, resulted in a slight reduction in accuracy.

\begin{table}[thb]\centering
    \resizebox{0.8\textwidth}{!}{
    \large
    \begin{tabular}{*{10}{c}}
        \toprule
        Start LR & Top-1 Accuracy(\%) & Top-5 Accuracy(\%) & Time per Epoch (s) \\
        \midrule
        1/4 & 73.73 & 93.11 & 112.99\\	
        1/8 & 73.78 & 92.75 & 113.22\\
        1/10 & $\mathbf{74.99}$ & $\mathbf{93.61}$ & 115.16\\
        1/16 & 74.50 & 93.12 & 115.59\\
        \bottomrule
    \end{tabular}
    }
    \caption{Comparison of results of different start learning rates on the Cifar-100 validation set with the ResNet18 backbones trained for 70 epochs. We set r = 0.1. The best results are bold.
    }
    \label{tab: Table 10}
\end{table}


\section{Limitation}
While FMDConv shows clear improvements in balancing speed and accuracy across various benchmarks, its deployment in highly resource-constrained environments may still face challenges due to the additional computational overhead introduced by the multi-attention mechanisms. Additionally, the model’s performance has been validated primarily on image classification tasks with ResNet architectures, leaving its efficacy on more complex tasks (such as object detection or segmentation) and other network architectures relatively unexplored. 

\section{Conclusion}\label{sec6}
In this paper, we introduced FMDConv, a novel dynamic convolution block designed to balance speed and accuracy. By leveraging three optimal attention mechanisms, FMDConv demonstrated superior performance on image classification benchmarks, making it suitable for resource-constrained environments. Our proposed metrics, the Inverse Efficiency Score (IES) and the Rate-Correct Score (RCS), effectively quantify the trade-offs between efficiency and accuracy. Future work will focus on extending FMDConv to more complex tasks, such as object detection, while further optimizing computational efficiency and exploring adaptive hyperparameter strategies for broader applicability.





\bibliographystyle{elsarticle-num-names} 
\bibliography{sn-bibliography}

\end{document}